\documentclass[letterpaper, 10 pt, conference]{ieeeconf}  
\makeatletter
\let\NAT@parse\undefined
\makeatother

\IEEEoverridecommandlockouts                              

\usepackage{graphics} 
\usepackage{epsfig} 
\usepackage{mathptmx} 
\usepackage{times} 
\usepackage{amsmath} 
\usepackage{amssymb}  
\usepackage{xcolor}
\usepackage{caption}
\usepackage{subcaption}
\usepackage{graphicx}
\usepackage[numbers,sort&compress]{natbib}

\title{\LARGE \bf
NEEDLE BIOPSY AND FIBER-OPTIC COMPATIBLE ROBOTIC INSERTION PLATFORM
}

\author{Fanxin Wang$^{1}$, Yikun Cheng$^{2}$, Chuyuan Tao$^{2}$, Rohit Bhargava$^{3}$, Thenkurussi Kesavadas$^{4}$ 
\thanks{$^{1}$Fanxin Wang, is with the Department of Mechatronics and Robotics, Xi'an Jiaotong-Liverpool University, 215123 China.
        {\tt\small Fanxin.Wang@xjtlu.edu.cn}}%
\thanks{$^{2}$Yikun Cheng and Chuyuan Tao are with the Department of Mechanical Science and Engineering, University of Illinois Urbana-Champaign, Champaign, IL 61801 USA. 
        {\tt\small  yikun2@illinois.edu, chuyuan2@illinois.edu}}%
\thanks{$^{3}$Rohit Bhargava are with Beckman Institute for Advanced Science and Technology, University of Illinois Urbana-Champaign, Champaign, IL 61801 USA.  
        {\tt\small sudiptam@illinois.edu, rxb@illinois.edu}}%
\thanks{$^{5}$Thenkurussi Kesavadas, is with the Division of Research and Economic Development, University at Albany - State University of New York, Albany, NY 12222 USA.  
        {\tt\small  tkesavadas@albany.edu}}%
}

\begin{document}

\maketitle
\thispagestyle{empty}
\pagestyle{empty}

\begin{abstract}
Tissue biopsy is the gold standard for diagnosing many diseases, involving the extraction of diseased tissue for histopathology analysis by expert pathologists. However, this procedure has two main limitations: 1) Manual sampling through tissue biopsy is prone to inaccuracies; 2) The extraction process is followed by a time-consuming pathology test. To address these limitations, we present a compact, accurate, and maneuverable robotic insertion platform to overcome the limitations in traditional histopathology. Our platform is capable of steering a variety of tools with different sizes, including needle for tissue extraction and optical fibers for vibrational spectroscopy applications. This system facilitates the guidance of end-effector to the tissue and assists surgeons in navigating to the biopsy target area for multi-modal diagnosis. In this paper, we outline the general concept of our device, followed by a detailed description of its mechanical design and control scheme. We conclude with the validation of the system through a series of tests, including positioning accuracy, admittance performance, and tool insertion efficacy.

\end{abstract}

\section{INTRODUCTION}
\begin{figure}
    \centering
    \includegraphics[width=0.97\linewidth]{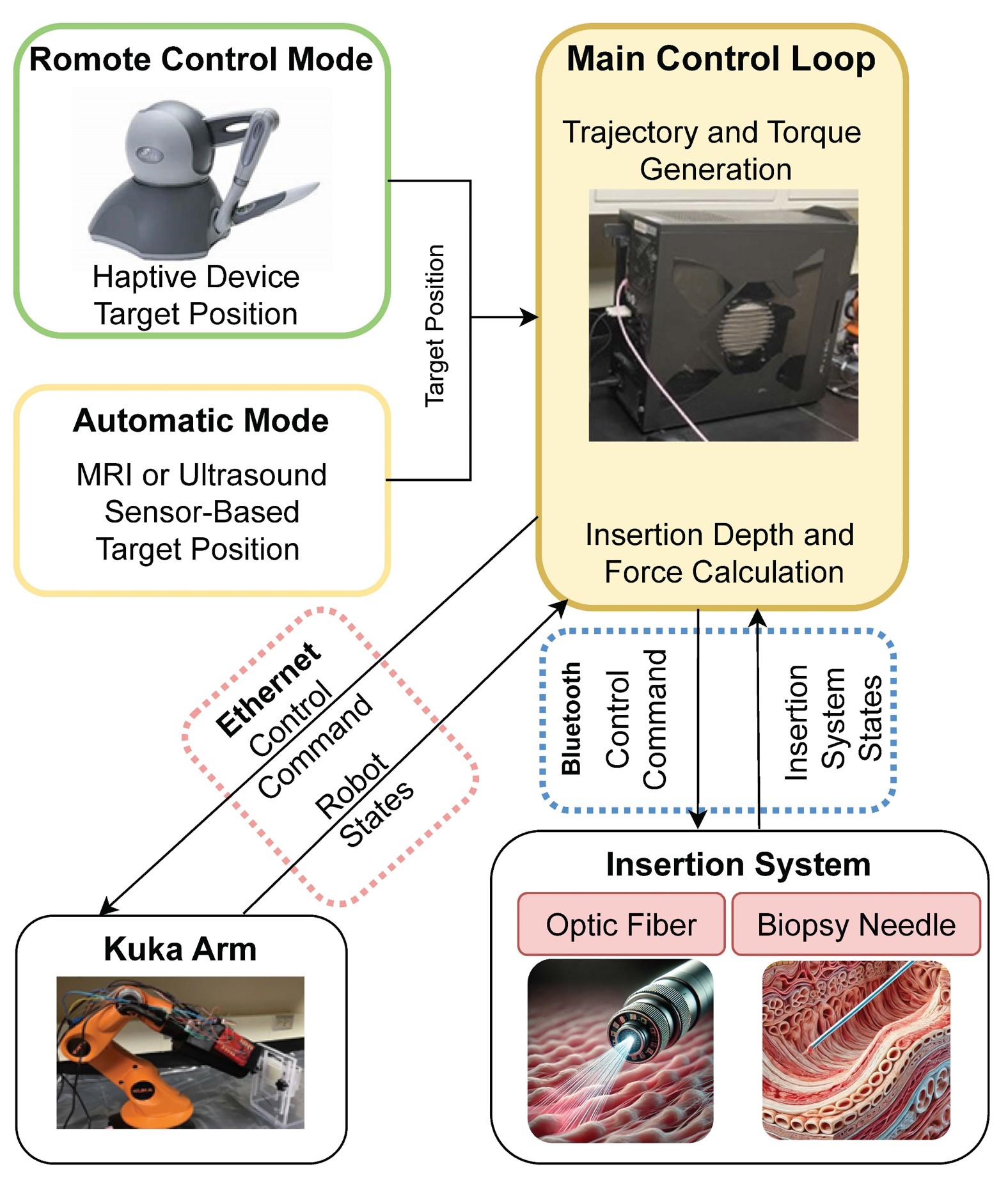}
    \caption{Overview of the proposed platform} 
    \label{fig:enter-label1}
\end{figure}
The approach for diagnosing cancers, as well as numerous other diseases, such as cardiomyopathy or hepatitis, encounters formidable limitations. Central to these limitations is the uncertainty surrounding the incremental benefits of tissue sampling and biopsy procedures \cite{c1}. Several factors contribute to this uncertainty. One critical issue is the potential for sampling errors, as biopsies may fail to include the specific affected areas within the region of interest. Additionally, inherent constraints on the amount of focal tissue that can be safely biopsied limit the depth of penetration, further complicating the diagnostic process. In response to these limitations, the medical field has explored advanced techniques to enhance diagnostic accuracy during biopsy procedures like electroanatomic mapping \cite{c2}, cardiac magnetic resonance imaging (CMR) \cite{c3}, and positron emission tomography \cite{c4}. These technologies have been utilized to guide the biopsy process more effectively, enabling the sampling of precisely affected myocardial regions. While these advanced techniques hold promise in improving the diagnostic accuracy of biopsy procedures, it's important to note that they are not yet widely accessible or routinely performed \cite{c5} or highly restricted in certain specific usage. Factors such as equipment availability, expertise required for interpretation, and cost constraints may limit their use in clinical practice. As such, efforts should continue to make these advanced diagnostic tools more accessible to clinicians,  potentially revolutionizing the way cardiomyopathy and other diseases are diagnosed and treated in a more robust and accurate way. 

Recent developments in robotic biopsy are making significant impacts on various usages in different disease diagnosis. It excels in performing precise and accurate needle insertions with the assistance of imaging systems \cite{c5}. Robotic biopsy procedures have been employed in diverse anatomical sites, such as bone \cite{c6}, lung \cite{c7}, breast \cite{c8}, brain or brain stem \cite{c9}, prostate \cite{c10}, and liver \cite{c11}. Robotic biopsy demonstrates its advantages in maintaining a straight-line trajectory, with guidance primarily relying on standard radiological images. In such situations, the robot serves a dual role: firstly, it combines diagnostic and intervention images through fusion, and secondly, it uses this image fusion to establish a direct path from a suitable external position to the intended target point. These types of interventions exemplify how robotics enhances precision by seamlessly integrating imaging with precise procedural guidance \cite{c5}. In 2018, researchers \cite{c12} pioneered the application of robotic assistance to address the patchy and unpredictable nature of infiltrating locations in traditional endomyocardial biopsy. They utilized a 150-cm bioptome with an alligator clamp attached to its proximal end, which was introduced through the Sensei Robotic System (Hansen Medical, Mountain View, CA). However, beyond robotic biopsy with conventional bioptome, there appears to be a  need to fill in rapid diagnosis, which is a gap between advanced vibrational spectroscopy technology to integrate with robotic platforms \cite{c5}.

The current diagnostic process (with or without robotic needle insertion) requires time-consuming pathology examination, and a majority of the biopsy specimens may be normal. Minimizing unnecessary tissue trauma during sampling is critical, as non-therapeutic injury to healthy tissue remains a key limitation of conventional biopsy practices \cite{c13}. The past two decades have seen the development of many in-vivo imaging techniques \cite{c14} that are based on optical spectroscopy. Former studies \citep{c15, c16} have underscored the practicality of digital histopathology in biopsy procedures by employing fiber-optic catheters, showcasing the method's capability to enhance diagnostic efficiency. These studies illustrate how digital techniques can transform traditional pathology by providing real-time, detailed tissue analysis. However, a notable limitation in these pioneering efforts is the lack of  robotic assistance, which restricts the precision and accuracy of needle insertions during biopsies. 

Our research focuses on developing a framework that guides end-effectors insertion efficiently and precisely through robotics manipulators to target specific regions of interest. The overview of proposed framework is depicted in Fig.1. The end-effector is capable of delivering conventional biopsy needle or optic fiber, which could be applied in the further study of combining vibrational spectroscopy. A 5 degree-of-freedom (DOF) supports the insertion system and delivers motion to the target tissue. It communicates with the central processor via Ethernet, exchanging control commands and robot state data. Additionally, a 3-DOF tool insertion system—comprising a standalone insertion module and a pitch control unit—interfaces with the central processor to relay control commands and system status with Bluetooth communication. The biopsy needle or the fiber-optic tool could be mounted on the tool insertion system with the spring-loaded roller mechanism. The tool insertion module is engineered to perform complex motions, including linear and rotational movements, in which optic fiber would explore through the tissue for molecular information. Fundamentally, our robotic insertion platform is adept at accurate tissue extraction using a standard biopsy needle. In advanced user scenarios, it can also be adapted for the insertion of optic fibers for vibrational spectroscopy, providing the potential to rapidly diagnosis in-vivo to revolutionize clinical practice. This can significantly reduce procedure times, alleviate patient anxiety, and lighten the workload for clinicians and pathologists. The primary focus of this paper is the development of this technology, presenting its design on tool compatibility and maneuverable control features, with a series of validation tests on a functional prototype of the steerable insertion system.

\begin{figure}
    \centering
    \vspace*{1.2mm}
    \includegraphics[width=0.85\linewidth]{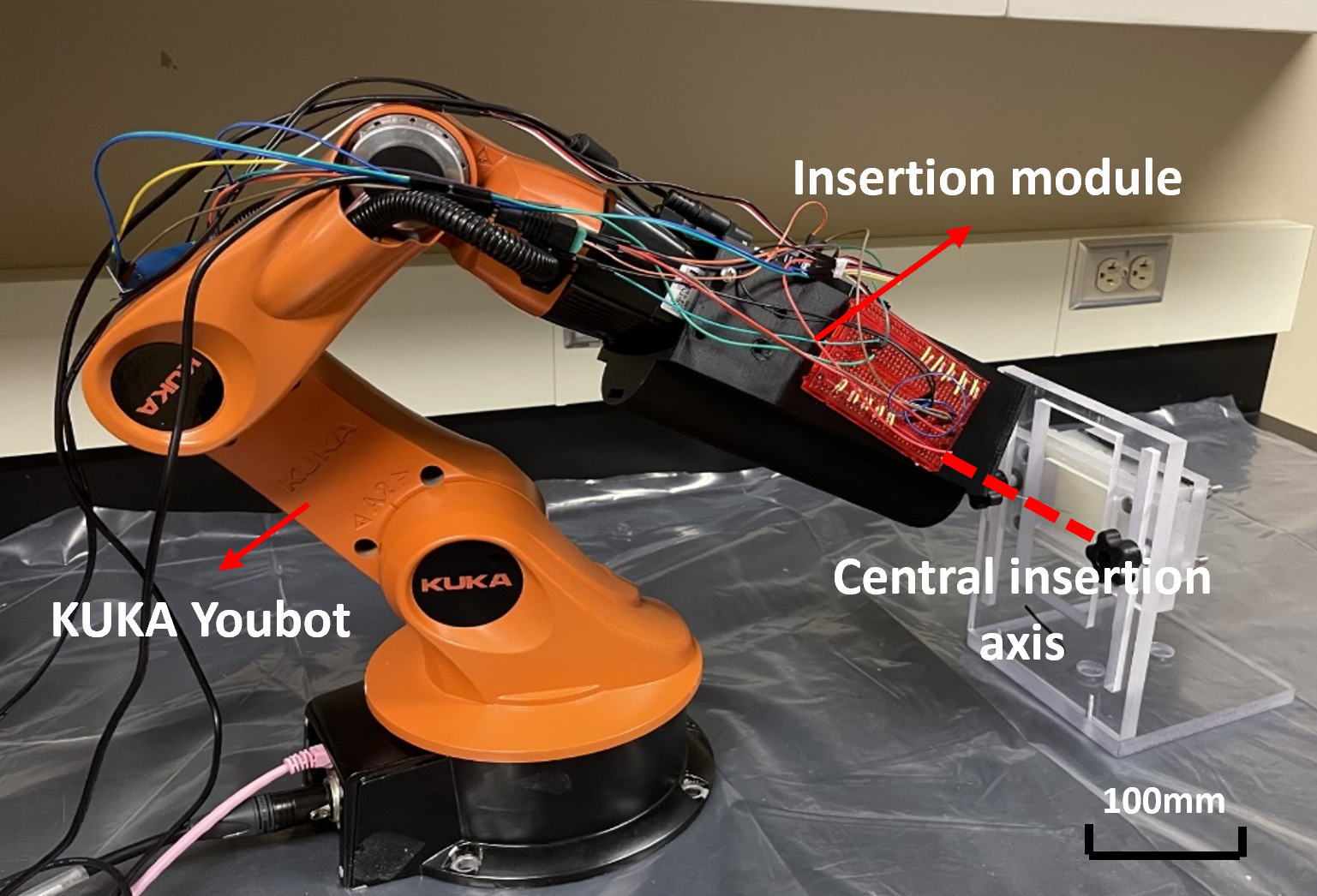}
    \label{fig:second}
    \caption{Real-world assembly of robotic insertion system}
    \label{fig:enter-label2}
\end{figure}
\vspace*{1mm}

\section{System and mechanical design}

\subsection{System Workflow and Overview}

Our proposed framework is a combination of exploiting tool catheters structures and robotic assistance for digital biopsy \cite{c17} and is described in detail in the following sections. Existing methods like ultrasound can be used to locate areas of interest for biopsy \cite{c18,c19,c20}. Based on the location within the anatomy of the organ, the insertion axis needs to be determined at the desired location that signifies the establishment of a directional reference within the anatomy \cite{c21,c22,c23}. By manipulating the orientation of the end-effector, it becomes possible to align it precisely with the insertion axis. The projected insertion tool is mounted on the insertion module, once the end-effector is perpendicular to target tissue, it will simultaneously translate and rotate about its axis in order to accurately sample tissue at different depths. This insertion procedure not only secures robust biopsy needle to reach the region of interest, it also enables the alternative tool tip -- needle for tissue extraction or optic fiber for exploring molecular information. 

\subsection{Platform Design Requirements}

The innovative principle of feed-force sensing in the proposed robotic system primarily ensures safety during in-situ needle-probe insertion \cite{c24}. Additionally, the system provides a generic steerable platform to ensure accurate access to desired location for in-situ biopsy or molecular information gathering. 

The above considerations lead to the following requirements for the design features: 

1). High dexterity with more than 6-DOF to provide the steerable platform with maneuverability; 

2). Strong insertion in programmable helical motion to collect/explore samples from the desired location, which requires at least 10N on the insertion \cite{c25}; 

3). Accommodation to a variety of tool sizes with user-defined mounting.

\subsection{Mechanical Design }

To achieve the goals stated above, we have the following mechanical designs details. The overall robotic platform consists of robotic arm supporting frame and insertion subsystem. The visual representation of the real-world assembled needle biopsy and fiber-optic compatible robotic insertion platform is illustrated in Fig.2, showcasing the integration of robotic supporting frame, tool insertion subsystem covered in end-effector holder and central insertion axis.

The utilization of a 5-DOF KUKA youBot arm provides a robust and versatile supporting frame for the steerable insertion system. This robotic frame is capable of precise and coordinated movements, offering the necessary dexterity and flexibility to manipulate the insertion effectively.

The insertion subsystem attached as the end-effector consists of two essential components: \textbf{standalone tool insertion module} and \textbf{pitch control unit}. To create a cohesive and functional system, the insertion subsystem is seamlessly mounted to the KUKA robot arm. This integration allows for precise coordination between the two systems. 

The standalone tool insertion module (shown in Fig.3(a)) is responsible for the actual insertion of the biopsy needle or optic fiber into the patient's tissue \cite{c17}. The standalone tool insertion module is primarily composed of a spring-loaded mechanism, slip-ring sensor modules, and two motors with driving gears. The tool is mounted using a clutching unit, which includes an actuation roller and a passive roller, as illustrated in Fig.3(b). The spring-loaded mechanism, which is connected to the passive roller, allows for user-defined clamping behavior and is designed to accommodate a variety of tool sizes. This flexibility is crucial for ensuring that the module can be used with different types of biopsy needles or optic fibers. The linear motion required for insertion is facilitated by the actuation roller, which is powered by motor $m_2$. This motor drives the roller, enabling the tool to be pushed into the tissue with precision. The rotational motion, essential for aligning the tool and for the optic fiber's ability to probe the tissue, is driven by motor $m_1$, which is connected to the floating clutching unit via a gearbox. Encoder data and force sensor data are transmitted to the controller using the slip-ring sensor modules. These sensors ensure that the data is continuously collected even as the tool rotates and feeds in to the tissue, providing real-time feedback to the system.
Also, the separation of motors is designed to achieve the precise translation and rotation at speeds that can be programmed according to the specific requirements of the insertion procedure. It allows for fine adjustments based on user input commands, ensuring that the tool, especially on needle insertion targeting sensitive tissues would follow the desired trajectory. 

While the standalone tool insertion module is adept at achieving precise translation and rotation, the tool insertion pitch angle also necessitates meticulous fine-tuning of the insertion axis. This fine-tuning is crucial to prevent imbalances in the supporting robotic frame placement, which could affect the accuracy and safety of the procedure. This feature not only enhances the precision of the system but also addresses safety concerns by ensuring that undesired needle insertions do not occur until the fine-tuning of the insertion axis is complete. By meticulously controlling the pitch, the system can accurately position the needle for biopsy, thereby minimizing the risk of damage to surrounding tissues and structures. 

\begin{figure}
    \centering
\centering
\begin{subfigure}{0.39\textwidth}
    \vspace*{2mm}
    \includegraphics[width=\textwidth]{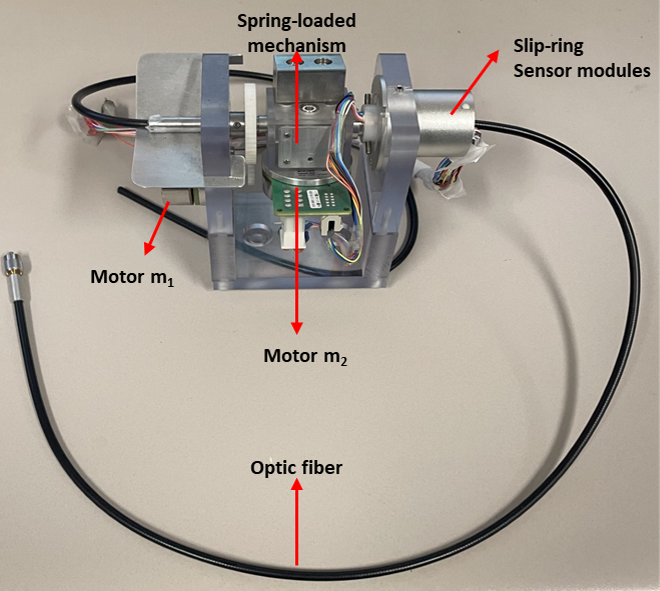}
    \caption{Real-world assembly of insertion module with fiber}
    \label{fig:first3}
\end{subfigure}
\hfill
\begin{subfigure}{0.37\textwidth}
    \includegraphics[width=\textwidth]{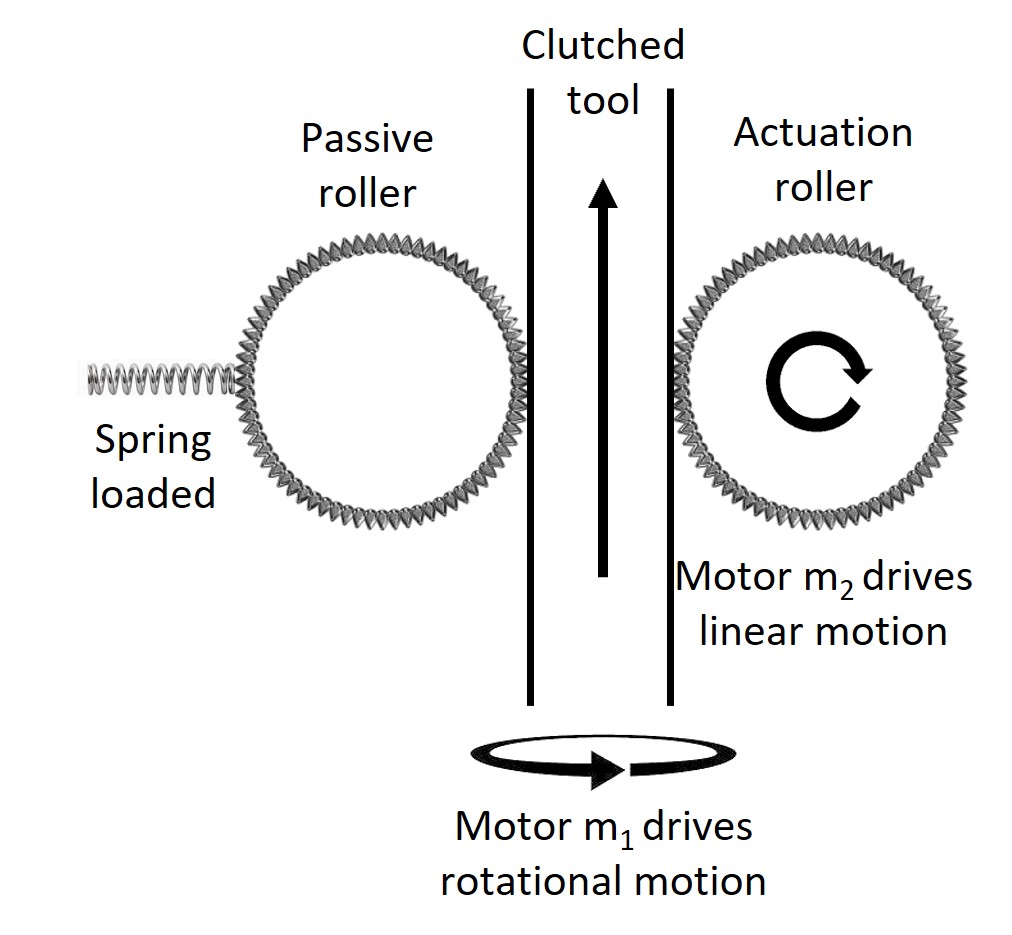}
    \caption{Illustration of insertion module mechanism}
    \label{fig:second3}
\end{subfigure}
    \caption{Standalone insertion module}
    \label{fig:enter-label3}
\end{figure}

\section{Robotic Platform Control Design}

The proposed semi-autonomous control for tool insertion is illustrated in the block scheme of Fig.4. 




\begin{figure}
    \centering
    \vspace*{2mm}
    \includegraphics[width=0.9\linewidth]{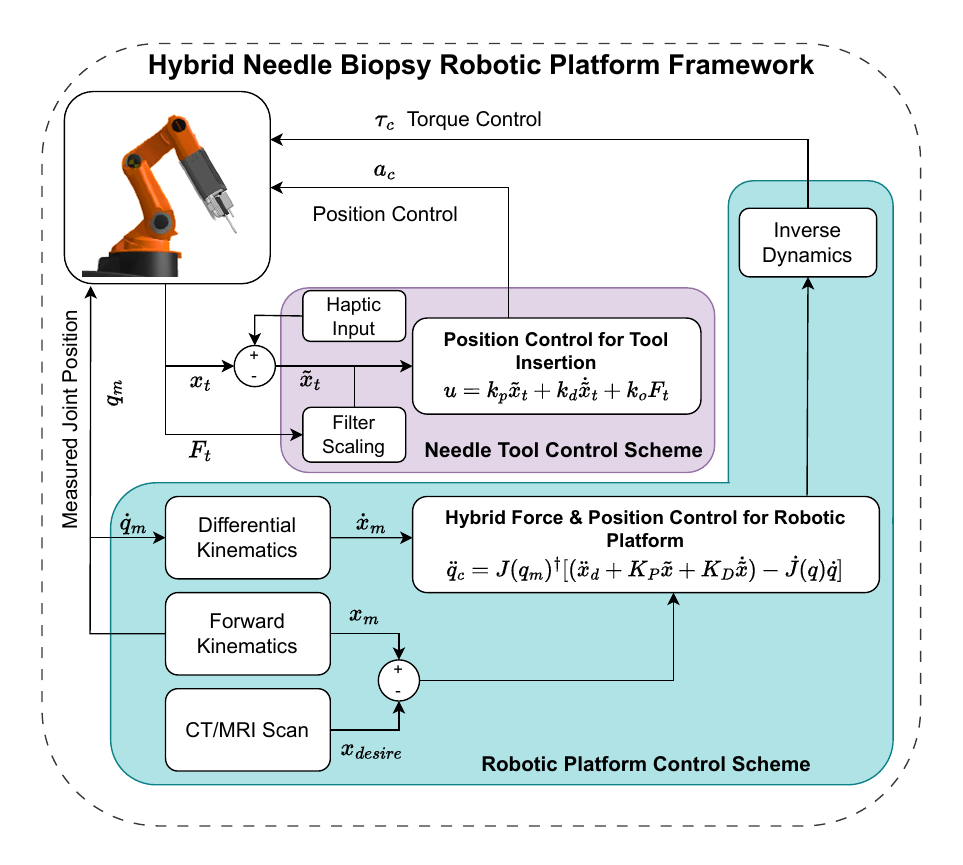}
    \caption{Control system diagram}
    \label{fig:enter-label4}
\end{figure}

\subsection{Control Design of Robotic Platform} 
The robotic platform dynamics and control law are stated as follows:
\begin{align}
    \tau_c &= M(q_m) \ddot{q}_c + C(q_m, \dot{q}_m) + G(q_m), \\
    \ddot{q}_c &= J(q_m)^\dagger \left[ \left( \ddot{x}_d + K_P \tilde{x} + K_D \dot{\tilde{x}} \right) - \dot{J}(q_m)\dot{q}_m \right]
\end{align}
where $M(q_m)$ represents the inertia matrix of the robot, $C(q_m, \dot{q_m})$ denotes the estimation of centrifugal and Coriolis effects, $G(q_m)$ is the gravity term, $\tau_c$ is the computed robotic command torque for each joint, and $q_m, \dot{q_m}$ represent the measured robot's joint position and angular velocity. $J(q_m)^\dagger$ is the pseudo-inverse of the robot's geometric Jacobian, and $\tilde{x} = [\tilde{p}, \tilde{\phi}]$ stands for the position error of the end-effector in the base reference frame and $\dot{\tilde{x}}$ is the relative speed error. 

The positioning trajectory from the robotic platform's initial condition to the target position is generated by a central processor within pre-recorded MRI/CT images. From this trajectory, we obtain the time-varying $x_d$, and we can calculate $\tilde{x}$ and $\tilde{\dot{x}}$ by subtracting the measured task-space position $x_m$ (using forward kinematics) and $\dot{x_m}$ (using differential kinematics). $\ddot{x}_d$ is the corresponding acceleration in the generated motion.

$K_P$ and $K_D$ are the proportional and derivative control gains, defined as $6 \times 6$ diagonal matrices in equations (3)(4).
\begin{align}
    K_P &= \text{diag}\{k_{p1}, k_{p2}, k_{p3}, k_{p4}, k_{p5}, k_{p6}\}, \\
    K_D &= \text{diag}\{k_{d1}, k_{d2}, k_{d3}, k_{d4}, k_{d5}, k_{d6}\}. 
\end{align}

Using equation (2), we obtain the desired joint acceleration, and we can then use inverse dynamics to calculate the commanded joint torques $\tau_c$.

\subsection{Admittance Control for Robotic Platform } 
Under ideal circumstance, the target position where the end-effector will be aligned with the insertion axis would be fixed if the patient does not move. But in real surgical environments, certain moves of the patients’ poses or the operation bed heading angle can lead to misalignment of the end-effector to the target insertion axis. Thus, manual operation of fine positioning is needed.
The goal of admittance control is to implement the behavior of the task space:
\begin{equation}
    F_{\text{ext}} = M\ddot{x} + B\dot{x} + Kx,
\end{equation}
where $\ddot{x} \in \mathbb{R}^6$ is the task-space configuration in a minimum set of coordinates, and $F_{\text{ext}}$ is an external force applied to the robot by the operator. 

We would compensate all internal forces including the friction torque and the gravity, dynamic terms in the joint torque equation, so that the external force would be the driving factor. 

 \subsection{Control Design of the End-effector} 
The end-effector operates semi-autonomously under the surgeon's control via a haptic device with force feedback. The tool tip's orientation is determined by the previously introduced control scheme, while the end-effector's control design primarily focuses on the insertion axis. During tool insertion, the measured resistance force varies depending on patient-specific conditions. To meet high-precision requirements, the end-effector must perform insertions under the surgeon's teleoperated guidance while incorporating real-time feedback.

In the semi-autonomous mode, the following control algorithm is introduced as:
\begin{equation}
    u = k_p \tilde{x}_t + k_d \dot{\tilde{x}}_t + k_o F_t,
\end{equation}
where $u$ is the calculated control output, $\tilde{x}_t$ is the 1 dimensional insertion axis position error, calculated by subtracting the scaled position $x_h$ from the haptic device input and the measured tool position. $\dot{\tilde{x}}_t$ is the 1 dimensional insertion axis velocity error calculated from the position error, providing a reference point for precise control. $F_t$ is the measured force along the insertion axis, reflecting the resistance force encountered during insertion. $k_p$, $k_d$, and $k_o$ are the tunable control gain factors for position, velocity, and force, respectively.

\section{Experiments and results}
The experiments are structured into two distinct parts to comprehensively evaluate the system's performance:
\\1.End-effector positioning within the hybrid robotic platform, assessing accuracy and admittance performance.
\\2.End-effector insertion control, testing tool insertion on different tissues and assessing needle displacement accuracy.

\subsection{Hybrid Robotic Platform Control Results with Admittance Performance}

To comprehensively assess the capabilities of the hybrid force/position control system, a series of pre-calculated trajectories were employed for evaluation. The first set of trajectories involved distinct sine waves executed separately along three axes with an amplitude of 0.05 meters, while the remaining axes and end-effector pointing direction remain constant. As shown in Fig.5, with mean errors in three axes tracking measuring less than 1 millimeter, and a standard error of tracking below 6\% (around 3 millimeters), these results provide concrete evidence of the system's consistent and reliable performance. These tight error margins indicate that the hybrid force/position control loop excels in maintaining trajectory fidelity, even when executing separate trajectory waves along distinct axes ensuring control and reliability needed for minimally invasive surgeries, patient safety, and overall surgical efficiency.

The end-effector's admittance response is shown in Fig.6 when subjected to external forces. A sequence of manual operation is applied randomly on joints to place the robotic supporting frame at a desired location (shown in red path). Between each set of manual operation, tests of self-holding is also delivered to check the stability of control. It is shown in Fig.6 that our platform has the reliable ability of to maintain the intended trajectory and achieve customized positioning even when faced with varying external forces. Supplementary demonstration videos provide a real-time view of the platform's performance in action.

\subsection{Position Control Result for Tool Insertion}
To comprehensively assess the position tracking accuracy along predefined trajectories and simultaneously capture the corresponding tissue reaction forces, we tracked the position (IR robot: IR-10$k\Omega$ linearity potentiometer) and reaction force, (ATI model: Nano 43 Transducer) with a biopsy cut needle (Bard instrument: 16-gauge/1.7mm) in the synthetic tissue samples with different types and thickness are shown in Fig.7(a). Supporting frame are shown in Fig.7(b) to hold the synthetic tissue samples in the desired position. 
Different setup for tissue samples are demonstrated as the following:
(1)	2mm skin-superficial tissue + 10mm fibrous tissue
(2)	2mm skin-superficial tissue + 15mm duct embedded tissue
(3)	4mm skin-superficial tissue + 10mm fibrous tissue
(4)	4mm skin-superficial tissue + 15mm duct embedded tissue
Continuous input signal of 1mm/s is commanded to the needle positioning. Relative force feedback and position response is recorded.

\begin{figure}
    \centering
    \vspace*{3mm}
    \includegraphics[width=0.75\linewidth]{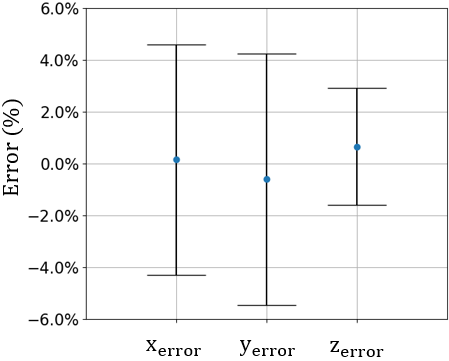}
    \caption{Robotic frame tracking error percentage result}
    \label{fig:enter-label5}
\end{figure}

\begin{figure}
    \centering
    \vspace*{3mm}
    \includegraphics[width=0.71\linewidth]{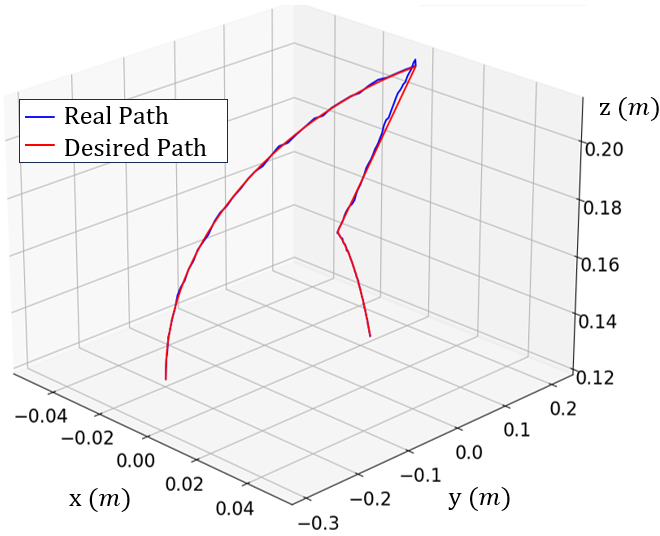}
    \caption{Admittance control result: Subjection to manual placement and automatic post-placement keeping.}
    \label{fig:enter-label6}
\end{figure}

\begin{figure}[!t]
     \centering
     \begin{subfigure}{0.27\textwidth}
         \centering
         \includegraphics[width=\textwidth]{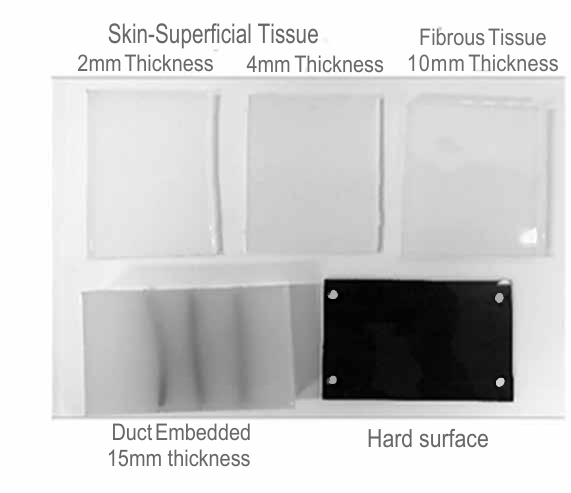}
         \caption{Synthetic tissue samples.}
         \label{fig_setup_a}
     \end{subfigure}
     \hspace{-2mm}
     \begin{subfigure}{0.205\textwidth}
         \centering
         \includegraphics[width=\textwidth]{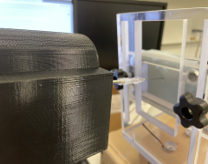}
         \vspace{2mm}
         \caption{Tissue supporting frame.}
         \label{fig_setup_b}
     \end{subfigure}
         \caption{Tool insertion experiment setup.}
        \label{fig3}
\end{figure}

\begin{figure}
    \centering
    \vspace*{3mm}
    \includegraphics[width=0.78\linewidth]{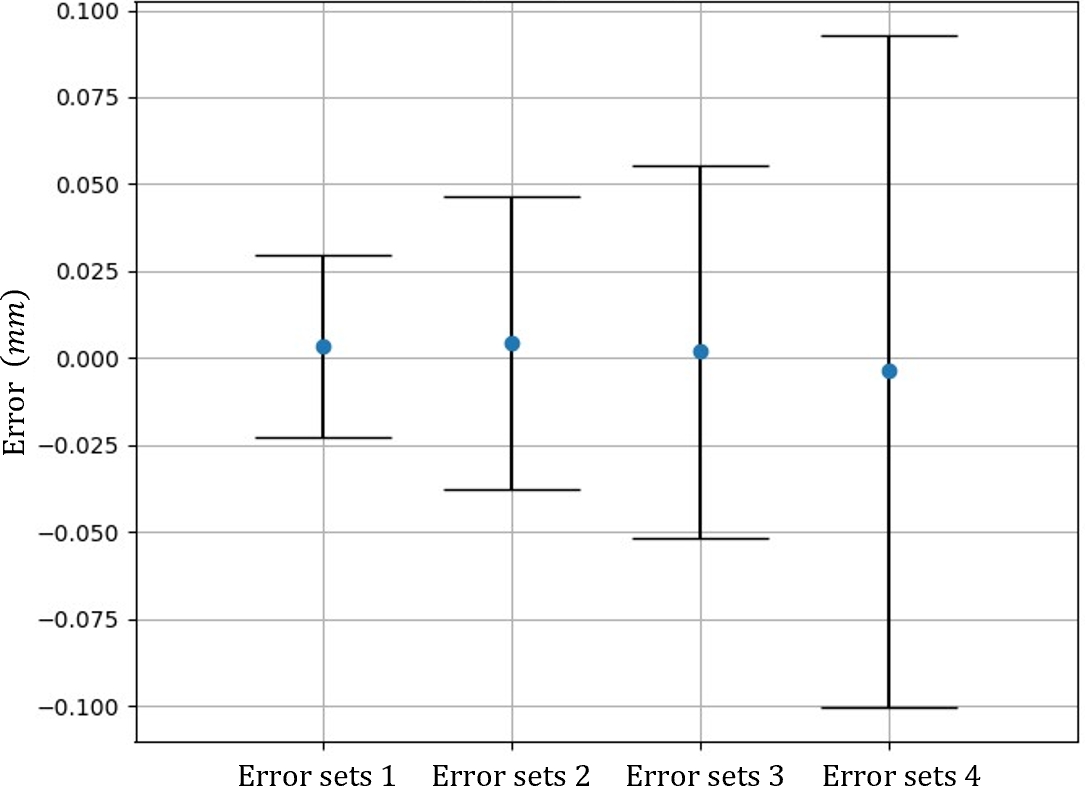}
    \caption{Error in needle insertions}
    \label{fig:enter-label7}
\end{figure}

A continuous input signal of 2mm/s is utilized to command needle positioning, demonstrating the system's control capabilities. As depicted in Fig.7, the evaluation across different tissue setups reveals that the needle insertion process encounters distinct penetration points and resistance forces, which are influenced by the specific tissue characteristics. In both scenarios examined, penetration is observed to occur at approximately 2mm of puncture, marked by a sudden drop in the reaction force. The subsequent resistance forces encountered during the insertion into heart and liver tissues are characterized by differences in tissue elasticity. This variation in resistance forces underscores the system's adaptability to tissue-specific properties, emphasizing the critical role of understanding tissue behavior in the field of medical robotics. This highlighting of the system's ability to manage the complexities of needle penetration is further researched as classification of tissue type \cite{c28}. The overall trajectory tracking error is illustrated in Fig.8, where needle insertions are carried out on different tissues for 10mm insertion, and the overall error percentage is less than 2\%.

\section{Conclusion and Future Work}

In this paper, a steerable robotic insertion system with needle-biopsy and fiber-optic compatibility is presented. The design achieves high dexterity to provide the steerable platform with accurate access to desired location, carries out strong and rapid incision to collect samples and is compatible with dexterous motion of optic fiber needed for label-free in situ optical spectroscopy.

There could be two major aspects for future development. Firstly, the current reference target position of tool insertion is fully relying on imaging system. While in real surgical situations, even with pre-recorded CT/MRI scans and ultrasonic guidance, accurately locating the target position of interest can be challenging. In our ongoing research, we are exploring methods to improve the accuracy of tool insertion by investigating the tissue properties. This includes learning techniques such as deep learning \cite{c26} or reinforcement learning \cite{c27} based on tissue feedback forces to determine target tissue properties. The potential advantages of integrating machine learning or reinforcement learning with tissue feedback force analysis hold great promise, and we expect these developments to significantly enhance the accuracy and efficiency of our steerable insertion system also with path planning \cite{c29} and collision avoidance \cite{c30}. Secondly, another area of future research is the mitigation of intrinsic vibrations caused by physiological processes such as breathing or heartbeats. These vibrations can compromise the accuracy of tool insertion, particularly through the incisions on body side. Developing strategies to counteract or compensate for such movements is crucial for maintaining tissue collection/exploration accuracy. Real-time motion tracking and predictive algorithms could be applied to ensure that the stability and precision despite the dynamic environment within the body.

\addtolength{\textheight}{-12cm}   




\section*{ACKNOWLEDGMENT}

We would like to thank Naveen Kumar Sankaran’s prior research and his design in standalone tool insertion module. We also acknowledge partial support from Mayo Clinic Foundation.

\addtolength{\textheight}{12cm}


\end{document}